\documentclass[10pt,twocolumn,letterpaper]{article}

\usepackage[pagenumbers]{wacv} 

\usepackage[dvipsnames]{xcolor}

\usepackage{graphicx}
\usepackage{amsmath}
\usepackage{amssymb}
\usepackage{booktabs}
\usepackage{caption}
\usepackage{graphicx}
\usepackage{multirow}
\usepackage{comment}
\usepackage{dsfont}
\usepackage{pifont}
\usepackage{dashrule}
\newcommand{\cmark}{\ding{51}}%
\newcommand{\xmark}{\ding{55}}%

\usepackage[accsupp]{axessibility}
\usepackage[pagebackref,breaklinks,colorlinks]{hyperref}

\usepackage[capitalize]{cleveref}
\crefname{section}{Sec.}{Secs.}
\Crefname{section}{Section}{Sections}
\Crefname{table}{Table}{Tables}
\crefname{table}{Tab.}{Tabs.}

\def\wacvPaperID{*****} 
\def\confName{WACV}
\def\confYear{2024}

\begin{document}

\title{A Survey on Multimodal Large Language Models for Autonomous Driving}


\author{%
  Can Cui$^{1}$\thanks{Equal contribution.}, Yunsheng Ma$^{1}$\footnotemark[1], Xu Cao$^{3,6}$\footnotemark[1], Wenqian Ye$^{4,6}$\footnotemark[1], Yang Zhou$^{5}$, Kaizhao Liang$^{7}$, Jintai Chen$^{3}$, \\
  Juanwu Lu$^{1}$, Zichong Yang$^{1}$, Kuei-Da Liao$^{8}$, Tianren Gao$^{7}$, Erlong Li$^{2}$, Kun Tang$^{2}$, Zhipeng Cao$^{2}$, \\
  Tong Zhou$^{2}$, Ao Liu$^{2}$, Xinrui Yan$^{2}$, Shuqi Mei$^{2}$, Jianguo Cao$^{6}$\thanks{Correspondence authors.}, Ziran Wang$^{1}$\footnotemark[2], Chao Zheng$^{2}$\footnotemark[2] \\
  $^{1}$ Purdue University, West Lafayette, IN, USA 47907 \\
  $^{2}$ Tencent T Lab, Beijing, China \\
  $^{3}$ University of Illinois Urbana-Champaign, Champaign, IL, USA 61801 \\
  $^{4}$ University of Virginia, Charlottesville, VA, USA 22903 \\ 
  $^{5}$ New York University, New York, NY, USA 11201 \\
  $^{6}$ PediaMed AI, Shenzhen, China \\
  $^{7}$ SambaNova Systems, Inc, Palo Alto, CA, USA 94303 \\
  $^{8}$ Objective, Inc, San Francisco, CA, USA 94110 \\
}

\maketitle
\begin{abstract}
With the emergence of Large Language Models (LLMs) and Vision Foundation Models (VFMs), multimodal AI systems benefiting from large models have the potential to equally perceive the real world, make decisions, and control tools as humans. In recent months, LLMs have shown widespread attention in autonomous driving and map systems. Despite its immense potential, there is still a lack of a comprehensive understanding of key challenges, opportunities, and future endeavors to apply in LLM driving systems. In this paper, we present a systematic investigation in this field. We first introduce the background of Multimodal Large Language Models (MLLMs), the multimodal models development using LLMs, and the history of autonomous driving. Then, we overview existing MLLM tools for driving, transportation, and map systems together with existing datasets and benchmarks. Moreover, we summarized the works in The 1st WACV Workshop on Large Language and Vision Models for Autonomous Driving (LLVM-AD), which is the first workshop of its kind regarding LLMs in autonomous driving. To further promote the development of this field, we also discuss several important problems regarding using MLLMs in autonomous driving systems that need to be solved by both academia and industry. Paper collection can be found at \href{https://github.com/IrohXu/Awesome-Multimodal-LLM-Autonomous-Driving}{Awesome-Multimodal-LLM-Autonomous-Driving}.
\end{abstract}

\section{Introduction}
\label{sec:intro}

\begin{figure}[!t]
    \centering
    \includegraphics[width=\linewidth]{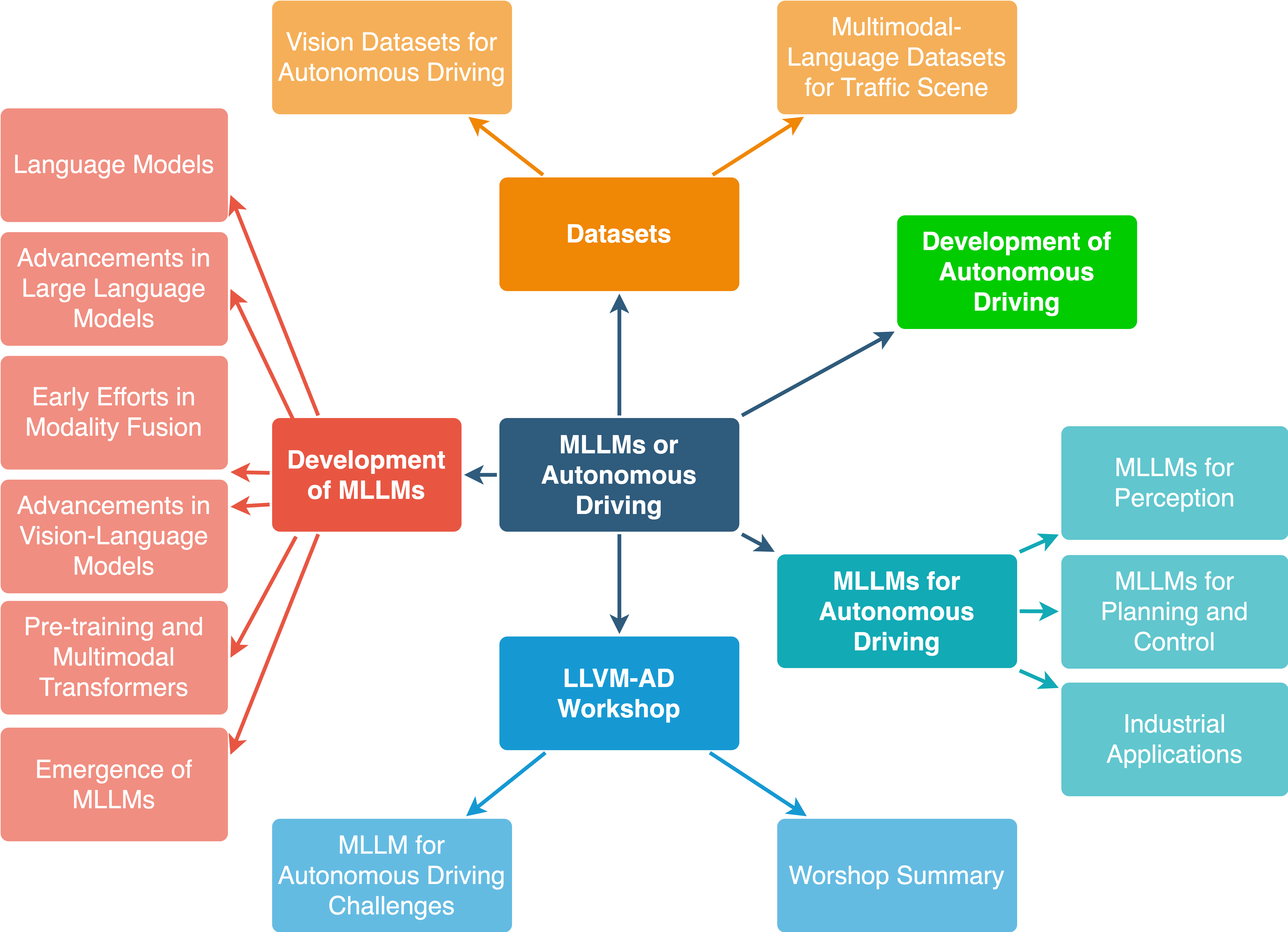}
    \caption{This survey paper focuses on the use of Multimodal Large Language Models (MLLMs) in the advancement of autonomous driving. The organization of the paper will delve into various aspects related to this topic.}
    \label{fig:overview}
\end{figure}

\begin{figure*}
    \centering
    \fbox{%
    \begin{minipage}{0.98\textwidth}
        \begin{minipage}{0.95\textwidth}
            \centering
            \begin{minipage}{0.65\textwidth}
                \paragraph{\color{black}Question / Prompt:\color{black}} \color{black} You are a professional self-driving agent and good at observing LiDAR point clouds and making driving decisions. I will show you some BEV point cloud images. Please use simple, clear, structured language to describe the road condition, surroundings, abnormal situation and driving strategy. \color{black}    
                
                \paragraph{\color{black}GPT-4V Answer:\color{black}} \color{gray} The BEV LiDAR image suggests a multi-lane road, but the exact number of lanes and presence of traffic are unclear. There are no visible obstacles, pedestrians, or significant environmental features in the immediate vicinity of the vehicle. The image lacks detail on road surface quality and surrounding signage or traffic lights. In this scenario, maintaining a steady speed while staying within the detected lanes, and being alert for unseen hazards is advisable. \color{black}
            \end{minipage}%
            \hfill
            \begin{minipage}{0.3\textwidth}
            \paragraph{\color{black}Input LiDAR Point Cloud:~\cite{tang2023thma}\color{black}}
                \centering
                \vspace{0.2cm}
                \includegraphics[width=0.96\linewidth]{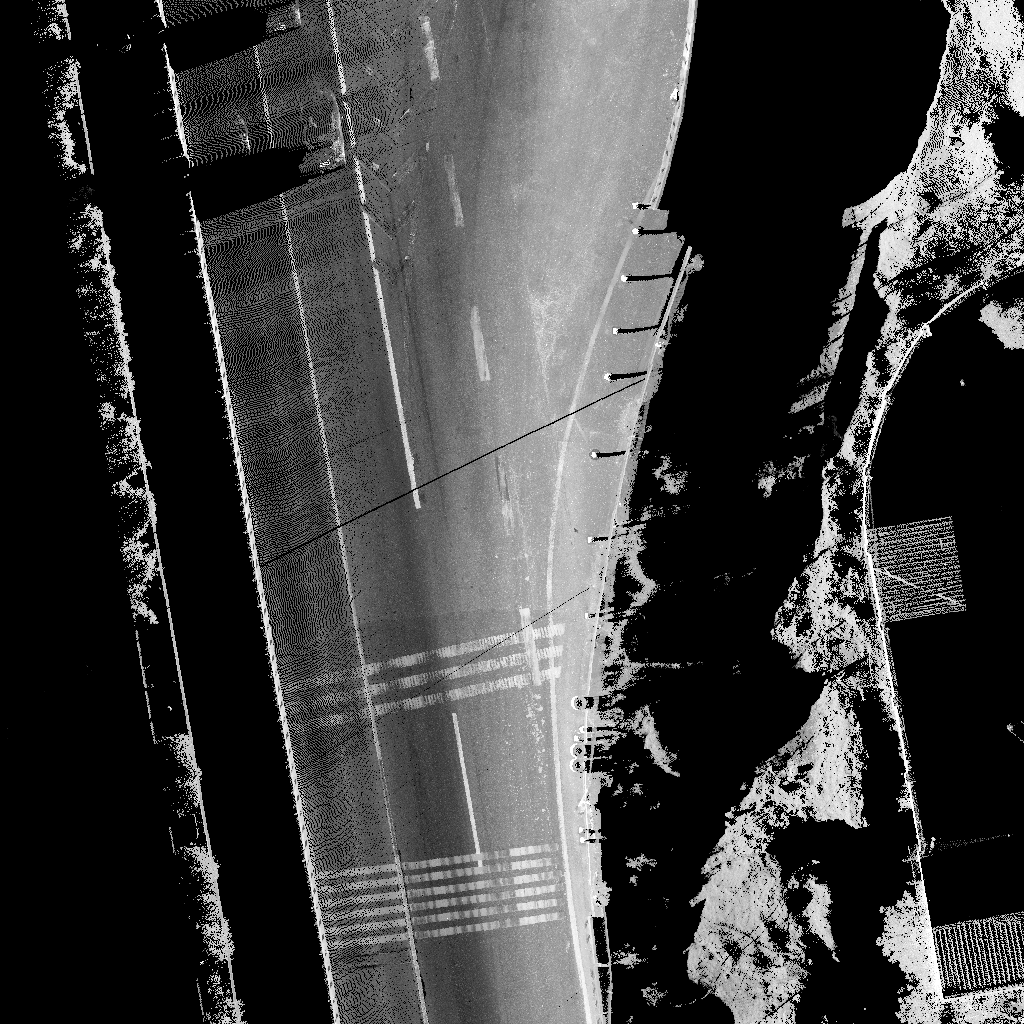}
                \vspace{0.2cm}
            \end{minipage}%
        \end{minipage}%

        \vspace{-0.05cm}
        \hdashrule[0.5ex]{0.95\linewidth}{1pt}{3pt}
        \vspace{-0.05cm}

        \begin{minipage}{0.95\textwidth}
            \centering
            \begin{minipage}{0.65\textwidth}
                \paragraph{\color{black}Question / Prompt:\color{black}} \color{black} As the AI system on an autonomous vehicle, your role is to analyze the driving context and make informed decisions. You will receive an image from the forward-facing camera of the ego vehicle. Please use concise and organized language to describe the current environment, any anomalies, and plan the next steps. \color{black}
                
                \paragraph{\color{black}GPT-4V Answer:\color{black}} \color{gray} Driving in an urban setting, the forward-facing camera captures a clear day with multi-story buildings lining both sides of a road, with some patches appearing wet or muddy. The road, populated with cars, motorcycles, and a three-wheeled vehicle. An anomaly to note is the three-wheeled vehicle on the left, which is carrying a heavy load that might affect its stability. Given the situation, the next steps involve maintaining a safe distance from this vehicle, monitoring motorcycles for sudden movements, scanning for traffic signals or signages, and adjusting speed to match the traffic flow. \color{black}
            \end{minipage}%
            \hfill
            \begin{minipage}{0.3\textwidth}
                \paragraph{\color{black}Input Driving Front View:\color{black}}
                \centering
                \vspace{0.2cm}
                \includegraphics[width=0.82\linewidth]{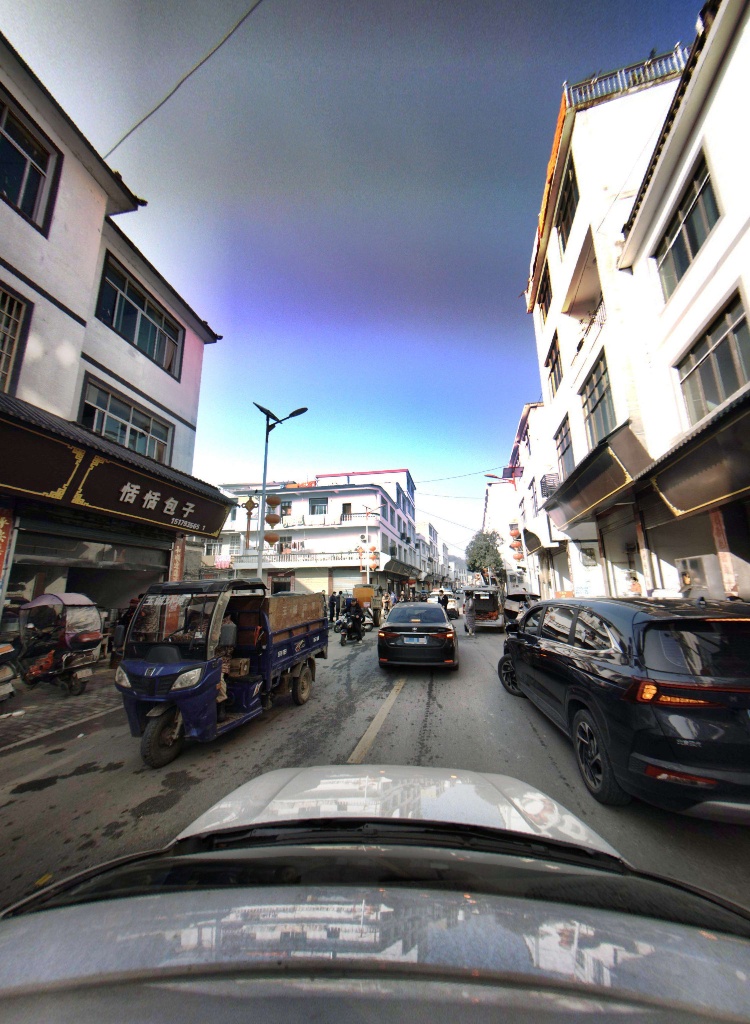}
                \vspace{0.2cm}
            \end{minipage}%
        \end{minipage}%

        \vspace{-0.05cm}
        \hdashrule[0.5ex]{0.95\linewidth}{1pt}{3pt}
        \vspace{-0.05cm}

        \begin{minipage}{0.95\textwidth}
        \centering
        \begin{minipage}{0.65\textwidth}
        \paragraph{\color{black}Question / Prompt:\color{black}} \color{black} Make a right lane change.\color{black}

        \paragraph{\color{black}GPT-4 Code Genration:\color{black}\vspace{1mm}}
        
        \centering
        \includegraphics[width=0.95\linewidth]{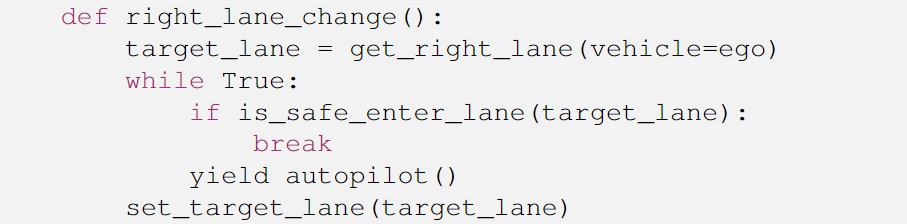}
        \end{minipage}%
        \hfill
        \begin{minipage}{0.3\textwidth}
            \paragraph{\color{black}Simulation~\cite{leurent_environment_2018}:\color{black}\vspace{1mm}}
            \centering
            \includegraphics[width=\linewidth]{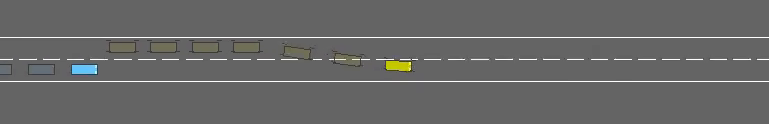}
        \end{minipage}%
        \end{minipage}
    \end{minipage}
    }
    \caption{Exploring GPT-4V~\cite{gpt4v2023} to understand driving scenes and make driving actions. Our findings reveal that while GPT-4V adeptly identifies scene components such as objects, it falls short in recognizing critical traffic elements like lane information. This underscores the significant challenges yet to be overcome in advancing multimodal language models for reliable autonomous vehicle navigation.}
    \label{fig:data_example}
\end{figure*}

Large Language Models (LLMs) have gained significant attention recently, showing remarkable potential in emulating human-like intelligence. The advancements have sparked a surge of enthusiasm in Multimodal Large Language Models (MLLMs)~\cite{yin2023survey}, which merge the sophisticated reasoning prowess of LLMs with image, video, and audio data. The modality alignment enables them to execute a diverse array of tasks with higher proficiency, including categorizing images, matching text to corresponding videos, and speech detection. Besides, \cite{vemprala_chatgpt_2023} proved that LLMs can deal with easy tasks within the robotics domain including basic logical, geometrical, and mathematical reasoning, to complex tasks such as aerial navigation, manipulation, and embodied agents. However, the integration of LLMs into the realm of transportation and autonomous vehicles is at a pioneering stage. Merging linguistic communication with multimodal sensory inputs like panoramic images, LiDAR point clouds, and driving actions could revolutionize the foundation models that govern current autonomous driving systems.

Recently, the emergence of more capable foundation models has made SAE L3 driving automation practicable~\cite{sae_j3016_2014}. However, the integration of multimodal LLMs in autonomous driving has not followed these advancements, and one natural question is, do LLM-based models like GPT-4, PaLM-2, and LLaMA-2 have the potential to enhance autonomous driving? Figure~\ref{fig:data_example} shows us a very good example. It is undeniable that integrating LLMs into the autonomous vehicle industry can bring a significant paradigm shift in vehicle intelligence, decision-making, and passenger interaction~\cite{cui_2024_WACV,cui_receive_2023}, offering a more user-centric, adaptable, and trustworthy future of transportation.

In the context of autonomous driving, LLMs will offer a transformative impact across crucial modules: perception, motion planning, and motion control~\cite{8944077}. In terms of perception, LLMs can harness external APIs to access real-time text-based information sources, such as HD maps, traffic reports, and weather updates, enabling the vehicle to attain a more comprehensive understanding of its surroundings~\cite{cui_receive_2023}. A good example is to improve the navigation in the vehicle-mounted maps. LLMs can process real-time traffic data to identify congested routes and suggest alternative paths, ultimately optimizing navigation for efficiency and safety~\cite{8967929}. For motion planning, LLMs play a role by utilizing their natural language understanding and reasoning ~\cite{mao_gpt-driver_2023}. They facilitate user-centric communication and enable passengers to express their intentions and preferences using everyday language. Additionally, LLMs also process textual data sources such as maps, traffic reports, and real-time information, and then make high-level decisions for optimized route planning~\cite{omama2023altpilot}. In the context of motion control, LLMs, firstly, enable the customization of controller parameters to align with driver preferences, achieving personalization in the driving experience~\cite{sha2023languagempc}. Additionally, LLMs can provide transparency by explaining each step of the motion control process.


MLLMs represent the next level of LLMs, bringing together the power of language understanding with the capability to process and integrate diverse data modalities~\cite{yin2023survey,driess_palm-e_2023}. Within the landscape of autonomous driving, the significance of MLLMs is huge and transformative. Vehicles equipped with MLLMs can deal with information from textual input with other features captured by onboard cameras and other sensors, offering easier learning of complex traffic scenes and driving behaviors. Beyond autonomous driving, MLLMs can also significantly enhance personalized human-vehicle interaction through voice communication and user preference analysis. In future SAE L4-L5 autonomous vehicles, passengers could communicate their requests while driving using language, gestures, or even gazes, with the MLLMs offering real-time in-cabin feedback by integrating visual displays or voice responses.

In our pursuit to bridge the domains of autonomous driving and advanced modeling, we co-organized the inaugural Workshop on Large Language and Vision Models for Autonomous Driving (\href{https://llvm-ad.github.io/}{LLVM-AD}) at the 2024 IEEE/CVF Winter Conference on Applications of Computer Vision (WACV). This event is designed to enhance collaboration between academic researchers and industry professionals, exploring the possibility and challenges of implementing multimodal large language models in the field of autonomous driving. LLVM-AD also launched a follow-up open-source real-world traffic language understanding dataset, catalyzing practical advancements. 

The main contributions of this paper are summarized as follows:

\begin{itemize}
    \item A brief overview of the background of current MLLMs and autonomous driving technologies is provided.
    \item The benefits of using LLMs and MLLMs in autonomous driving are outlined, highlighting their roles and current works in perception, motion planning, motion control, and recently declared industry applications.
    \item Datasets relevant to autonomous driving are summarized, with an emphasis on driving language datasets for traffic scenes.
    \item The accepted papers from the WACV \href{https://llvm-ad.github.io/}{LLVM-AD} Workshop are reviewed, providing insights into future directions of LLMs and MLLMs in autonomous driving.
\end{itemize}

As Figure~\ref{fig:overview} shows, our survey paper aims to provide a comprehensive overview of MLLMs for autonomous driving and discuss growing trends, and future directions. The following two sections provide a brief description of the developmental history of autonomous driving and MLLMs separately. Section~\ref{sec:mllm4ad} presents current published works about MLLMs for autonomous driving in perception, motion planning, and motion control. Section~\ref{sec:dataset} introduces related autonomous driving industry applications utilizing MLLMs. In the last three sections, we summarize the papers in the 1st WACV LLVM-AD workshop and discuss potential research directions for LLMs and MLLMs for autonomous driving.

\begin{figure*}[t!]
    \centering
    \includegraphics[width=\textwidth]{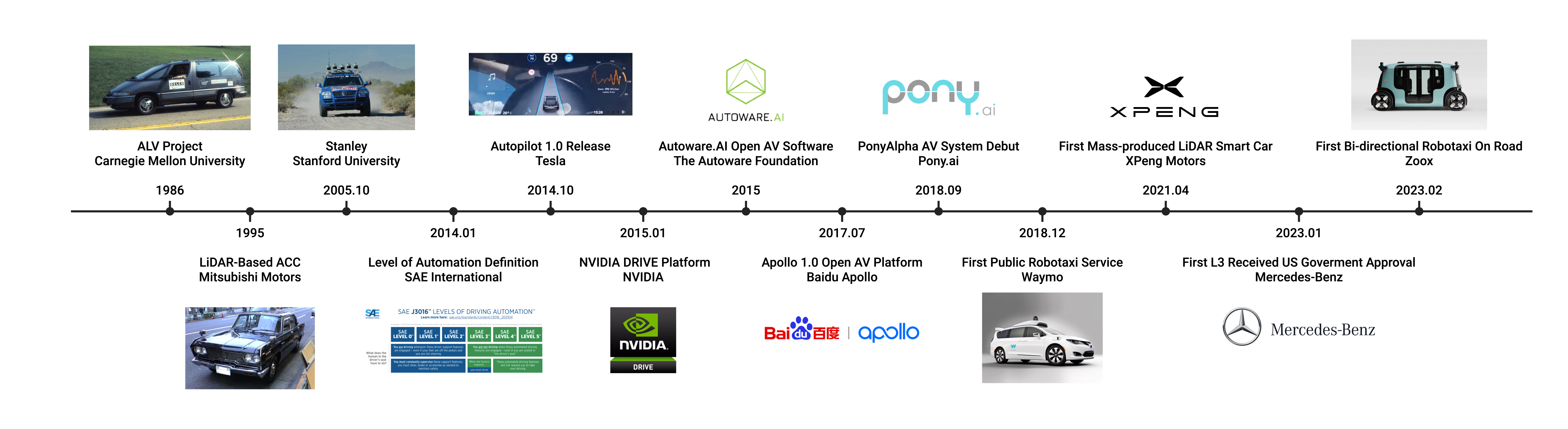}
    \caption{The figure outlines the chronological development of autonomous driving technology.
    It begins with representative early exploration and advancements like the ALV Project by Carnegie Mellon University~\cite{kanade_autonomous_1986, robot_hall_of_fame}, Mitsubishi Debonair the first to offer LiDAR-based ADAS system~\cite{nabhan2020models}, and winner of 2005 DARPA Grand Challenge Stanley by Stanford University~\cite{Stanley}. It then showcases recent achievements after the introduction of a standardized level of automation~\cite{sae_j3016_2014} and rapid progress in Deep Neural Networks. Autonomous driving platform-wise, various open source and commercialized software solutions are introduced, such as Tesla Autopilot~\cite{tesla_motors_manual}, NVIDIA DRIVE, Autoware.AI~\cite{autoware_1,autoware_2}, Baidu Apollo~\cite{apolloautoapollo_2023}, and PonyAlpha~\cite{ponyai}. Regulatory and service-wise, autonomous driving technology are receiving increasing government acceptance and public acknowledgment, with numerous companies receiving permits to operate autonomous driving vehicles on public roads in designated regions while more vehicles with autonomous driving capabilities are being mass-produced~\cite{group_mercedes-benz_2023}. Overall, it demonstrates the evolution and increasing sophistication of AD systems over several decades.}
    \label{fig:av_timeline}
\end{figure*}

\section{Development of Autonomous Driving}

The quest for autonomous driving has been a progressive journey, marked by a continuous interplay between visionary aspirations and technological capabilities. The first wave of comprehensive research on autonomous driving started in the late 20th century. For example, the Autonomous Land Vehicle (ALV) project launched by Carnegie Mellon University utilized sensor readings from stereo cameras, sonars, and the ERIM laser scanner to perform tasks like lane keeping and obstacle avoidance~\cite{pomerleau_alvinn1988, kanade_autonomous_1986}. However, these researches were constrained by limited sensor accuracy and computation capabilities.

The last two decades have seen rapid improvements in autonomous driving systems. A classification system published by the Society of Automotive Engineers (SAE) in 2014 defined six levels of autonomous driving systems~\cite{sae_j3016_2014}. The classification method has now been widely acknowledged and illustrated important milestones for the research and development progress. The introduction of Deep Neural Networks (DNNs) has also played a significant role ~\cite{kuutti_dnnsurvey_2021, Grigorescu_dnn_survey_2020}. Backed by deep learning, computer vision has been crucial for interpreting complex driving environments, offering state-of-the-art solutions for problems such as object detection, scene understanding, and vehicle localization ~\cite{qi_2017_pointnet, lang_2019_pointpillars ,CGV-079}. Deep Reinforcement Learning (DRL) has additionally played a pivotal role in enhancing the control strategies of autonomous vehicles, refining motion planning, and decision-making processes to adapt to dynamic and uncertain driving conditions ~\cite{drive_in_a_day, deep_rl_survey, waymo_rl_survey, cai_dq-gat_2022},. Moreover, sensor accuracy and computation power improvements allow larger models with more accurate results to be run on the vehicle. With such improvements, More L1 to L2 level Advanced Driver Assistance Systems (ADAS) like lane centering and adaptive cruise control are now available on everyday vehicles ~\cite{bengler_ADAS_overview, chen_iv_milestones_2023}. Companies like Waymo, Zoox, Cruise, and Baidu are also rolling out Robotaxis with Level 3 or higher autonomy. Nevertheless, such autonomous systems still fail in many driving edge cases such as extreme weather, bad lighting conditions, or rare situations ~\cite{cui_safety_review_2019}.

Inspired by current limitations, part of the research on autonomous driving is now focusing on addressing the safety of autonomous systems and enhancing the safety of autonomous systems ~\cite{ekim_safety_survey}. As Deep Neural Networks are often considered black boxes, trustworthy AI aims at making the system more reliable, explainable, and verifiable. For example, generating adversarial safety-critical scenarios for training autonomous driving systems such that the system is more capable of handling cases with low probability ~\cite{abey_adversarial_2019, ding_safety_2021}. Another way to improve the overall safety is through vehicle-to-infrastructure and vehicle-to-vehicle communication. With information from nearby instances, the system will have improved robustness and can receive early warnings ~\cite{ni_v2x_2020, liu_v2ip2i_2018}. Meanwhile, as Large Language Models show their powerful reasoning and scene-understanding capability, research is being conducted to utilize them to improve the safety and overall performance of the autonomous driving system.

\begin{figure*}[t!]
    \centering
    \includegraphics[width=\textwidth]{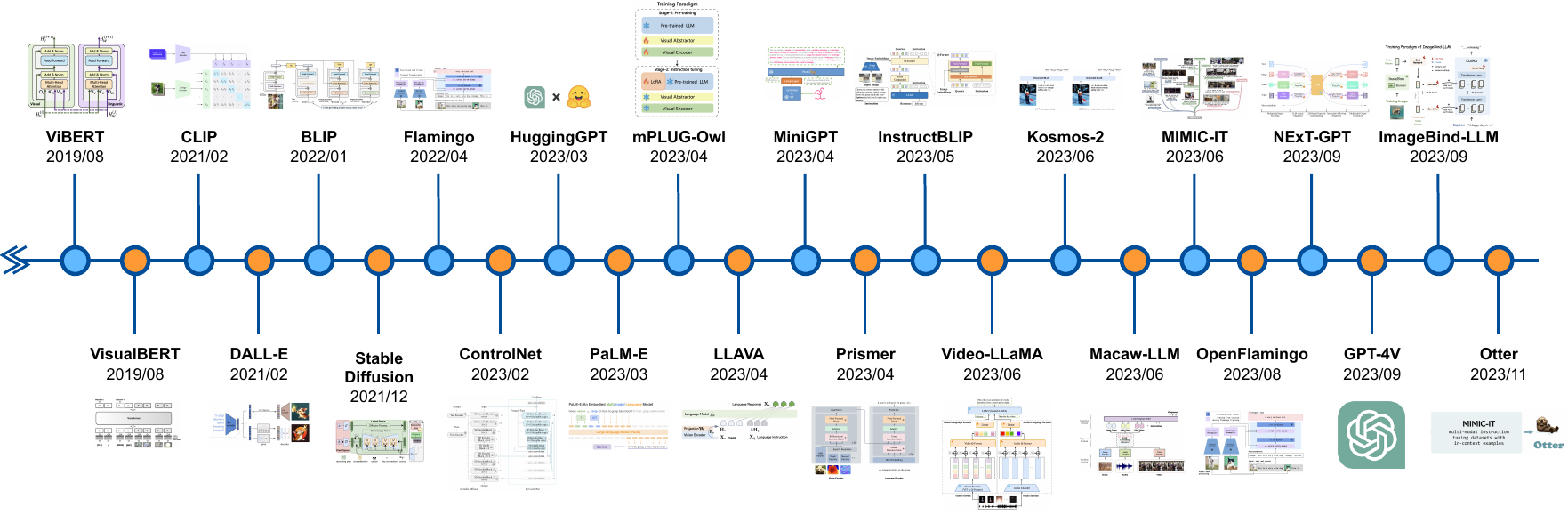}
    \caption{A timeline of recent advancements in Multimodal Large Language Models (MLLMs).}
    \label{fig:mllm_timeline}
\end{figure*}

\section{Development of Multimodal Language Models}

\subsection{Development of Language Models}
The development of language models has been a journey marked by significant breakthroughs. Since the early 1960s, many linguists, most renowned Noam Chomsky, attempted to model natural languages~\cite{chomsky2014aspects}. Early efforts focused mainly on rule-based approaches~\cite{bar1960present, oettinger1960automatic, holt1960man}. However, in the late 1980s and early 1990s, the spotlight shifted onto statistic models, such as N-gram ~\cite{brown1992class},  hidden Markov models~\cite{fine1998hierarchical}, which relied on counting the frequency of words and sequences in text data. The 2000s witnessed the introduction of neural networks into natural language modeling. Recurrent Neural Networks (RNNs) ~\cite{schuster1997bidirectional} and Long Short-Term Memory (LSTM) networks ~\cite{hochreiter1997long} were used for various NLP tasks.

Despite their potential, early neural models had limitations in capturing long-range dependencies and struggled with complex language tasks. In 2013, Tomas Mikolov and his team at Google introduced Word2Vec~\cite{mikolov2013efficient}, a groundbreaking technique for representing words as dense vectors, providing a better understanding of semantic relationships between words. This laid down the foundation for the rise of deep learning ~\cite{sutskever2014sequence, chung2014empirical}, which eventually led to the pivotal work, Attention is all you need ~\cite{vaswani2017attention}, which kick-started the new era of large language models.\cite{devlin2018bert, raffel2020exploring, radford2019language, brown_language_2020, chowdhery_palm_2022}.

\subsection{Advancements in Large Language Models}
LLMs are a category of Transformer-based language models known for their extensive number of parameters, often numbering in the hundreds of billions. These models are trained on vast amounts of internet data, which enables them to perform a wide range of language tasks, primarily through text generation. Some well-known examples of LLMs include GPT-3~\cite{brown_language_2020}, PaLM~\cite{chowdhery_palm_2022}, LLaMA~\cite{touvron_llama_2023}, and GPT-4~\cite{openai_gpt-4_2023}. One of the most notable characteristics of LLMs is their emergent abilities, such as in-context learning (ICL)~\cite{brown_language_2020}, instruction following~\cite{ouyang_training_2022}, and reasoning with chain-of-thought (CoT)~\cite{wei_chain--thought_2022}.

There is a growing area of research that utilizes LLMs to develop autonomous agents with human-like capabilities. These agents leverage the extensive knowledge stored in pre-trained LLMs to create coherent action plans and executable policies~\cite{liang_code_2023,wang_voyager_2023,ahn_as_2022,huang_language_2022,huang_voxposer_2023,driess_palm-e_2023}. Embodied language models~\cite{driess_palm-e_2023} directly integrate real-world sensor data with language models, establishing a direct connection between words and perceptual information. Voyager~\cite{wang_voyager_2023} introduces lifelong learning by incorporating three main components: an automatic curriculum that promotes exploration, a skill library to store and retrieve complex behaviors, and an iterative prompting mechanism to generate executable code for embodied control. Voxposer~\cite{huang_voxposer_2023} utilizes LLMs to generate robot trajectories for a wide range of manipulation tasks, guided by open-ended instructions and objects.

In parallel with these advancements, the use of LLMs in the field of autonomous driving is gaining momentum. Recent research~\cite{fu_2024_WACV, jin_surrealdriver_2023} has investigated the application of LLMs to comprehend driving environments. These studies have demonstrated the impressive ability of LLMs to handle complex scenarios by converting visual information into text representation, enabling LLMs to interpret the surrounding world. Similarly, in RRR~\cite{cui_receive_2023}, authors propose a human-centric autonomous driving framework that breaks down user commands into a series of intermediate reasoning steps, accompanied by a detailed list of action descriptions to accomplish the objective.

\subsection{Early Efforts in Modality Fusion}
Over the past few decades, the fusion of various modalities such as vision, language, video, and audio has been a key objective in artificial intelligence (AI). Initial efforts in this domain focused on simple tasks, such as image or video captioning and text-based image retrieval, which were mostly rule-based and relied on hand-crafted features. A classic example of early AI problems in the 1970s and 1980s was the "Blocks World"~\cite{slaney2001blocks}, where the goal was to rearrange colored blocks on a table based on textual instructions. This early attempt bridged vision (understanding block configurations) with language (interpreting and executing instructions), even though it was not based on deep learning.

\subsection{Advancements in Vision-Language Models}
In the following years, the field of multimodal models saw significant advancements. Over the last decade, the advent of deep learning has revolutionized approaches to visual-language tasks. Convolutional Neural Networks (CNNs)~\cite{krizhevsky2012imagenet} became the de facto standard for image and video processing, while Recurrent Neural Networks (RNNs)~\cite{schuster1997bidirectional, hochreiter1997long} emerged as the go-to models for processing sequential data, such as natural languages. During this period, popular tasks included image and video captioning, which involves generating descriptive sentences for images and videos, and visual question answering (VQA), where models answer questions related to visual data. Typical vision-language models employed joint embeddings, with image features (processed by CNNs) and text features (processed by RNNs or Transformers~\cite{vaswani2017attention}) mapped to a shared semantic space to facilitate multimodal learning~\cite{karpathy2015deep,mao2014deep,vinyals2015show,antol2015vqa}. Beyond vision and language, researchers also proposed models for other modalities, such as audio, speech, and 3D data. For instance, Mroueh et al. (2015) developed a deep multimodal learning model for audio-visual speech recognition that utilizes CNNs for visual data and RNNs for audio data~\cite{mroueh2015deep}. Arandjelović and Zisserman (2017) explored the relationship between visual and auditory data by developing a model that learns shared representations from unlabeled videos, using CNNs for both image and audio processing~\cite{arandjelovic2017look}. Furthermore, Qi et al. (2016) introduced models that process 3D data, including point clouds, for object classification tasks, employing CNNs to learn representations from volumetric data and multiple 2D views of 3D objects~\cite{qi2016volumetric}. These works highlight the potential of multimodal learning in capturing complex relationships between different types of data, leading to richer and more accurate representations.

\subsection{Pre-Training and Multimodal Transformers}

Building on this momentum, the field of multimodal models has continued to evolve, with researchers exploring the potential of pre-training multimodal models on extensive datasets before fine-tuning them on specific tasks. This approach has resulted in significant performance improvements across a range of applications. Inspired by the success of pre-trained NLP models like BERT~\cite{devlin2018bert}, T5~\cite{raffel2020exploring}, and GPTs~\cite{radford2018improving, brown_language_2020}, researchers developed multimodal Transformers that can process cross-modality inputs such as text, image, audio, pointcloud~\cite{huang2022masked,georgescu2023audiovisual,guo2023point}. Notable examples of visual-language models include CLIP~\cite{radford_learning_2021}, ViLBERT~\cite{lu2019vilbert}, VisualBERT~\cite{li2019visualbert}, SimVLM~\cite{wang2021simvlm}, BLIP-2~\cite{li2023blip} and Flamingo~\cite{alayrac_flamingo_2022}, which were pre-trained on large-scale cross-modal datasets comprising images and languages. Other works have explored the use of multimodal models for tasks such as video understanding~\cite{zhu2020actbert}, audio-visual scene understanding~\cite{alayrac2020self}, and even 3D data processing~\cite{han2020occuseg}. Pre-training allows the models to align different modalities and enhance the representation learning ability of the model encoder. By doing so, these models aim to create systems that can generalize across tasks without the need for task-specific training data. Furthermore, the evolution of multimodal models has also given rise to new and exciting possibilities. For instance, DALL-E~\cite{ramesh2021zero} extends the GPT-3 architecture to generate images from textual descriptions, Stable Diffusion~\cite{rombach2022high} and ControlNet~\cite{zhang2023adding} utilized CLIP and UNet-based diffusion model to generate images controlled by text prompt. They showcase the potential for using multimodal models in many application scenarios such as healthcare~\cite{liang2023pie}, civil engineering~\cite{ploennigs2023ai}, robotics~\cite{kang2023imagined} and, art~\cite{ko2023large}.

\subsection{Emergence of Multimodal Large Language Models}
Recently, MLLMs have emerged as a significant area of research. These models leverage the power of LLMs, such as ChatGPT~\cite{openai_chatgpt_2023}, InstructGPT~\cite{ouyang_training_2022}, FLAN~\cite{wei2022finetuned, chung_scaling_2022}, and OPT-IML~\cite{iyer2023optiml} to perform tasks across multiple modalities such as text and images. They exhibit surprising emergent capabilities, such as writing stories based on images and performing OCR-free math reasoning, which are rare in traditional methods. This suggests a potential path to artificial general intelligence. Key techniques and applications in MLLMs include Multimodal Instruction Tuning, which tunes the model to follow instructions across different modalities~\cite{zhu2023minigpt4, liu2023visual, ye2023mplugowl}; Multimodal In-Context Learning, which allows the model to learn from the context of multimodal data~\cite{yang2023mmreact,lu2023chameleon, gupta2023visual, dong2022survey, lu2022fantastically}; Multimodal Chain of Thought, which enables the model to maintain a chain of thought across different modalities~\cite{ge2023chain, zhang2023multimodal, rose2023visual, himakunthala2023lets}; and LLM-Aided Visual Reasoning (LAVR), which uses LLMs to aid in visual reasoning tasks~\cite{wang2023caption, zhang2023prompt, zhu2022pointclip, lu2023chameleon, gupta2023visual, shen2023hugginggpt, wu2023visual, yang2023mmreact}. MLLMs are more in line with the way humans perceive the world, offering a more user-friendly interface and supporting a larger spectrum of tasks compared to LLMs. The recent progress of MLLMs has been ignited by the development of GPT-4V~\cite{gpt4v2023}, which, despite not having an open multimodal interface, has shown amazing capabilities. The research community has made significant efforts to develop capable and open-sourced MLLMs, exhibiting surprising practical capabilities.

\begin{table*}[!t]
    \centering
    \resizebox{0.98\linewidth}{!}{
    \begin{tabular}{lcp{1.8cm}p{2cm}p{1.5cm}cp{1.2cm}p{1.5cm}}
        \hline
        Model & Year & Backbone & Task & Modality & Learning & Input & Output\\
        \hline
        Driving with LLMs~\cite{chen_driving_2023} & 2023 & LLaMA & Perception Control & Vector Language & FT & Vector Query & Response \qquad  Actions\\Talk2BEV~\cite{dewangan_talk2bev_2023}& 2023 & Flan5XXL Vicuna-13b & Perception Planning & Vision Language & ICL & Image Query & Response\\
        GAIA-1~\cite{hu_gaia-1_2023} & 2023 & - & Planning & Vision Language & PT & Video Prompt & Video \\
        LMaZP~\cite{huang_language_2022} & 2022 & GPT-3  Codex & Planning & Language & ICL  & Text & Plan \\
        Dilu~\cite{wen2023dilu} & 2023 & GPT-3.5  GPT-4 & Planning Control & Language & ICL & Text & Action\\
        DaYS~\cite{cui_2024_WACV} & 2023 & GPT-4 & Planning & Language & ICL & Text & Code \\
        RRR~\cite{cui_receive_2023} & 2023 & GPT-4 & Planning Control & Language & ICL & Text & Action \\
        DlaH~\cite{fu_drive_2023} & 2023 &GPT-3.5 & Planning Control & Language & ICL & Text & Action  \\
        GPT-Driver~\cite{mao_gpt-driver_2023} &2023 & GPT-3.5 & Planning & Vision Language & ICL & Text & Trajectory \\
        SurrealDriver~\cite{jin_surrealdriver_2023} & 2023 & GPT-4 & Planning Control & Language & ICL & Text & Text \qquad Action  \\
        LanguageMPC~\cite{sha2023languagempc} & 2023 & GPT-3.5 & Planning & Language & ICL & Text & Action\\
        DriveGPT4~\cite{xu_drivegpt4_2023} & 2023 & Llama 2 & Planning Control & Vision Language & ICL & Image Text Action & Text \qquad Action  \\
        \hline
    \end{tabular}
    }
    \caption{Summary of recent research on MLLMs for autonomous driving. The main backbone for current models are LLaMA~\cite{touvron_llama_2023-1}, Llama 2~\cite{touvron_llama_2023}, GPT-3.5~\cite{openai_chatgpt_2023}, GPT-4~\cite{openai_gpt-4_2023}, Flan5XXL~\cite{chung_scaling_2022}, Vicuna-13b~\cite{the_vicuna_team_vicuna_2023}. FT, ICL and PT refer to fine-tuning, in-context learning and pretrained respectively.}
    \label{tab:model}
\end{table*}

\section{Multimodal Language Models for Autonomous Driving}
\label{sec:mllm4ad}

In the autonomous driving industry, MLLMs have the potential to understand traffic scenes, improve the decision-making process for driving, and revolutionize the interaction between humans and vehicles. These models are trained on vast amounts of traffic scene data, allowing them to extract valuable information from different sources like maps, videos, and traffic regulations. As a result, they can enhance a vehicle's navigation and planning, ensuring both safety and efficiency. Additionally, they can adapt to changing road conditions with a level of understanding that closely resembles human intuition.

\subsection{Multimodal Language Models for Perception}

Traditional perception systems are often limited in their ability to recognize only a specific set of predefined object categories. This restricts their adaptability and requires the cumbersome process of collecting and annotating new data to recognize different visual concepts. As a result, their generality and usefulness are undermined. In contrast, a new paradigm is emerging that involves learning from raw textual descriptions and various modalities, providing a richer source of supervision.

Multimodal Large Language Models (MLLMs) have gained significant interest due to their proficiency in analyzing non-textual data like images and point clouds through text analysis~\cite{zhang_video-llama_2023,alayrac_flamingo_2022,radford_learning_2021,tsimpoukelli_multimodal_2021}. These advancements have greatly improved zero-shot and few-shot image classification~\cite{radford_learning_2021,p_proto-clip_2023}, segmentation~\cite{luddecke2022image, kirillov2023segment}, and object detection~\cite{minderer2022simple}.

Pioneering models like CLIP~\cite{radford_learning_2021} have shown that training to match images with captions can effectively create image representations from scratch. Building on this, Liu et al. introduced LLaMa~\cite{liu2023visual}, which combines a vision encoder with an LLM to enhance the understanding of both visual and linguistic concepts. Zhang et al. further extended this work with Video-LLaMa~\cite{zhang_video-llama_2023}, enabling MLLMs to process visual and auditory information from videos. This represents a significant advancement in machine perception by integrating linguistic and visual modalities.

Furthermore, researchers have explored the use of vectorized visual embeddings to equip MLLMs with environmental perception capabilities, particularly in autonomous driving scenarios. DriveGPT4~\cite{xu_drivegpt4_2023} interprets video inputs to generate driving-related textual responses. HiLM-D~\cite{ding_hilm-d_2023} focuses on incorporating high-resolution details into MLLMs, improving hazard identification and intention prediction. Similarly, Talk2BEV~\cite{dewangan_talk2bev_2023} leverages pre-trained image-language models to combine Bird's Eye View (BEV) maps with linguistic context, enabling visuo-linguistic reasoning in autonomous vehicles.

At the same time, progress in autonomous driving is not limited to discriminative perception models; generative models are also gaining popularity. One example is the Generative AI for Autonomy model (GAIA-1), which generates realistic driving scenarios by integrating video, text, and action inputs. This generative world model can anticipate various potential outcomes based on the vehicle's maneuvers, showcasing the sophistication of generative models in adapting to the changing dynamics of the real world~\cite{hu_gaia-1_2023}. Similarly, UniSim~\cite{yang_learning_2023} aims to replicate real-world interactions by combining diverse datasets, including objects, scenes, actions, motions, language, and motor controls, into a unified video generation framework. Moreover, the Waymo Open Sim Agents Challenge (WOSAC)~\cite{montali_waymo_2023,gulino_waymax_2023} is the first public challenge to develop simulations with realistic and interactive agents.



\subsection{Multimodal Language Models for Planning and Control}
The use of language in planning and control tasks has a longstanding history in robotics, dating back to the use of lexical parsing in natural language for early demonstrations of human-robot interaction~\cite{winograd_procedures_1971}, and it has been widely studied being used in the robotics area. There exists comprehensive review works on this topic~\cite{tellex_robots_2020,luketina_survey_2019}. It has been well-established that language acts as a valuable interface for non-experts to communicate with robots~\cite{kollar_toward_2010}. Moreover, the ability of robotic systems to generalize to new tasks through language-based control has been demonstrated in various works~\cite{ahn_as_2022, jang_bc-z_2022}. Achieving specific planning or control tasks or policies, including model-based~\cite{nair_learning_2021,andreas_learning_2017,sharma_correcting_2022}, imitation learning~\cite{lynch_language_2021, shridhar_cliport_2021}, and reinforcement learning~\cite{jiang_language_2019,goyal_pixl2r_2020,misra_mapping_2017}, has been extensively explored. 

Due to the significant ability in zero-shot learning~\cite{tong_zero-shot_2022}, in-context learning~\cite{min_rethinking_2022} and reasoning~\cite{wei_chain--thought_2022}, many works showed that LLMs could enable reasoning of planning~\cite{wang_voyager_2023,pmlr-v205-shah23b} and perceiving the environment with textual description~\cite{10161317} to develop user in the loop robotics~\cite{vemprala_chatgpt_2023}. \cite{kojima_large_2022} broke down natural language commands into sequences of executable actions through a combination of text completion and semantic translation to control the robot. SayCan~\cite{ahn_as_2022} utilized weighted LLMs to produce reasonable actions and control robots while \cite{huang_inner_2022} uses environmental feedback, LLMs can develop an inner monologue, enhancing their capacity to engage in more comprehensive processing within robotic control scenarios. Socratic Models~\cite{zeng_socratic_2022} employs visual language models to replace perceptual information within the language prompts used for robot action generation.~\cite{liang_code_2023} introduces an approach that uses LLMs to directly generate policy code for robots to do control tasks, specify feedback loops, and write low-level control primitives. 

In autonomous driving, LLMs could serve as the bridge to support human-machine interactions. For general purposes, LLMs can be task-agnostic planners. In \cite{huang_language_2022}, the authors discovered that pre-trained LLMs contain actionable knowledge for coherent and executable action plans without additional training. Huang et al.~\cite{huang_voxposer_2023} proposed the use of LLMs for converting arbitrary natural language commands or task descriptions into specific and detail-listed objectives and constraints. \cite{wen2023dilu} proposed integrating LLMs as decision decoders to generate action sequences following chain-of-thoughts prompting in autonomous vehicles. In \cite{cui_2024_WACV}, authors showcased that LLMs can decompose arbitrary commands from drivers to a set of intermediate phases with a detailed list of descriptions of actions to achieve the objective. 

Meanwhile, it is essential to enhance the safety and explainable of autonomous driving. The multimodal language model provides the potential to comprehend its surroundings and the transparency of the decision process. \cite{kim2018textual} showed that video-to-text models can help generate textual explanations of the environment aligned with downstream controllers. Deruyttere et al.~\cite{deruyttere_talk2car_2019} compared baseline models and showed that LLMs can identify specific objects in the surroundings that are related to the commands or descriptions in natural language. For the explainability of the model, Xu et al.~\cite{xu_drivegpt4_2023} proposed to integrate LLMs to generate explanations along with the planned actions. In~\cite{cui_2024_WACV}, the authors proposed a framework where LLMs can provide descriptions of how they perceive and react to environmental factors, such as weather and traffic conditions.

Furthermore, the LLMs in autonomous driving can also facilitate the fine-tuning of controller parameters, aligning them with the driver's preferences and thus resulting in a better driving experience. \cite{sha2023languagempc} integrates LLMs into low-level controllers through guided parameter matrix adaptation.

Besides the development of LLMs, great progress has also been witnessed in MLLMs. The MLLMs have the potential to serve as a general and safe planner model for autonomous driving. The ability to process and fuse visual signals such as images enhanced navigation tasks by combining visual cues and linguistic instructions \cite{kamath_new_2023,ku_room-across-room_2020}. Interoperability challenges have historically been an issue for autonomous planning processes \cite{gohel_explainable_2021,chib_recent_2023}.  However, recent advancements in addressing interoperability challenges in autonomous planning have leveraged the impressive reasoning capabilities of MLLMs during the planning phases of autonomous driving \cite{fu_2024_WACV,chen_driving_2023}. In one notable approach, Chen et al. \cite{chen_driving_2023} integrated vectorized object-level 2D scene representations into a pre-trained LLM with adapters, enabling direct interpretation and comprehensive reasoning about various driving scenarios. Additionally, Fu et al. \cite{fu_2024_WACV} employed LLMs for reasoning and translated this reasoning into actionable driving behaviors, showing the versatility of LLMs in enhancing autonomous driving planning. Additionally, GPT-Driver \cite{mao_gpt-driver_2023} reformulated motion planning as a language modeling problem and utilized LLM to describe highly precise trajectory coordinates and its internal decision-making process in natural language in motion planning. SurrealDriver \cite{jin_surrealdriver_2023} simulated MLLM-based generative driver agents that can perceive complex traffic scenarios and generate corresponding driving maneuvers. \cite{keysan2023text} investigated the utilization of textual descriptions along with pre-trained language encoders for motion prediction in autonomous driving.

\subsection{Industrial Applications}
The integration of MLLMs in the autonomous driving industry has been developed by several significant initiatives. Wayve introduces LINGO-1, which enhances the learning and explainability of foundational driving models by integrating vision, language, and action ~\cite{wayve_lingo-1_2023}. They also developed GAIA-1, a generative world model for realistic driving scenario generation, offering fine-grained control over vehicle behavior and scene features~\cite{hu_gaia-1_2023}.

Tencent T Lab generated traffic, map, and driving-related context from their HD map AI system~\cite{tang2023thma}, creating MAPLM, a large map and traffic scene dataset for scene understanding. 

Waymo's contribution, MotionLM, improved motion prediction in multi-agent environments. By conceptualizing continuous trajectories as discrete motion tokens, it transfers multi-agent motion prediction to a language modeling task~\cite{seff2023motionlm}. This approach transforms the dynamic interaction of road agents into a manageable sequence-to-sequence prediction problem.

Research from the Bosch Center focuses on using natural language for enhanced scene understanding and predicting future behaviors of surrounding traffic ~\cite{keysan2023text}. Meanwhile, researchers from the Hong Kong University of Science and Technology and Huawei Noah's Ark Lab have leveraged MLLMs to integrate various autonomous driving tasks, including risk object localization and intention and suggestion prediction from videos~\cite{ding_hilm-d_2023}.

These developments in industry illustrate the expanding role of MLLMs in enhancing the capabilities and functionalities of autonomous driving systems, marking a significant improvement in vehicle intelligence and situational awareness.

\section{Datasets and Benchmarks}
\label{sec:dataset}

\subsection{Vision Datasets for Autonomous Driving}
Publicly available datasets have played a crucial role in advancing autonomous driving technologies. \cref{tab:vision-dataset} provides a comprehensive overview of the latest representative datasets for autonomous driving. In the past, datasets mainly focused on 2D annotations, like bounding boxes and masks, primarily for RGB camera images~\cite{tusimple,pan2018spatial}. However, achieving autonomous driving capabilities that can match human performance requires precise perception and localization in the 3D environment. Unfortunately, extracting depth information from purely 2D images poses significant challenges.

To enable robust 3D perception or mapping, researchers have created many multimodal datasets. These datasets include not only camera images but also data from 3D sensors like radar and LiDAR. An influential example in this field is the KITTI dataset~\cite{geiger_are_2012}, which provides multimodal sensor data, including front-facing stereo cameras and LiDAR. KITTI also includes annotations of 3D boxes and covers tasks such as 3D object detection, tracking, stereo, and optical flow. Subsequently, NuScenes~\cite{caesar_nuscenes_2020} and the Waymo Open dataset~\cite{sun_scalability_2020} have emerged as representative multimodal datasets. These datasets set new standards by offering a large number of scenes. These datasets represent a significant advancement in the availability of large data for advancing research in autonomous driving.

\subsection{Multimodal-Language Datasets for Traffic Scene}
Several pioneering studies have explored language-guided visual understanding in driving scenarios. These studies either enhance existing datasets with additional textual information or create new datasets independently. The former category includes works such as Talk2Car~\cite{deruyttere_talk2car_2019}, nuScenes-QA~\cite{qian2023nuscenes}, DriveLM~\cite{drivelm2023}, and NuPrompt~\cite{wu2023language}. Among these, Talk2Car~\cite{deruyttere_talk2car_2019} stands out as the first object referral dataset, which contains natural language commands for autonomous vehicles. On the other hand, datasets like BDD-X~\cite{kim2018textual} and DRAMA~\cite{malla2023drama} were independently created. DRAMA~\cite{malla2023drama} specifically focuses on video and object-level inquiries regarding driving hazards and associated objects. This dataset aims to enable visual captioning through free-form language descriptions and uses both closed and open-ended responses to multi-tiered questions. It allows for the evaluation of various visual captioning abilities in driving contexts.

Despite the advancements in language comprehension in traffic scenes with MLLMs, their performance is still far below the human level. This is because traffic data-text pairs contain diverse modalities, such as 3D point clouds, panoramic 2D imagery, high-definition map data, and traffic regulations. These elements significantly differ from conventional domain contexts and question-answer pairs, highlighting the unique challenges of deploying MLLMs in that autonomous driving context. The datasets mentioned above are limited in terms of scale and quality, which hinders efforts to fully address these emerging challenges.
\begin{table}[!t]
    \centering
    \resizebox{\linewidth}{!}{
    \begin{tabular}{l|cccccccccc}
        \hline
        Dataset & Year & RGB & LiDAR & Text & Map\\
        \hline
        KITTI~\cite{geiger_are_2012} & 2012 & 15K & 15K & \xmark & \xmark\\
        nuScenes~\cite{caesar_nuscenes_2020} & 2019 & 1.4M & 400K & \cmark & \cmark\\
        Argo1~\cite{chang_argoverse_2019} & 2019 & 107K & 22K & \xmark & \cmark\\
        Waymo Open~\cite{sun_scalability_2020} & 2019 & 1M & 200K & \xmark & \cmark\\
        Argo2~\cite{wilson_argoverse_2021} & 2021 & 5.4M & 6M & \xmark & \cmark\\
        V2V4Real~\cite{xu_v2v4real_2023} & 2023 & 40K & 20K & \xmark & \cmark\\
        \hline
    \end{tabular}
    }
    \caption{Comparison of representative autonomous driving datasets.}
    \label{tab:language-dataset}
\end{table}

\begin{table*}[!htbp]
\small
\centering
\begin{tabular}{lcccccccc}
\toprule
\multirow{2}{*}{\textbf{Dataset}} & \multirow{2}{*}{\textbf{Year}} & \multirow{2}{*}{\textbf{QA}} & \multirow{2}{*}{\textbf{Caption}} & \multirow{2}{*}{\textbf{Scenario}} & \multirow{2}{*}{\textbf{Text}} & \multicolumn{3}{c}{\textbf{Modality}} \\ \cmidrule(lr){7-9} 
&&&&&& \textbf{Image.} & \textbf{Point Cloud.} & \textbf{Map Info.} \\ 
\midrule
BDD-X~\cite{kim2018textual} & 2018 & \xmark & \cmark & 7K & 26K & \cmark & \xmark & \xmark \\
Talk2Car~\cite{deruyttere_talk2car_2019} & 2019 & \xmark & \cmark  & 34K & 12K & \cmark & \xmark & \xmark \\
DRAMA~\cite{malla2023drama} & 2023 & \xmark & \cmark  &  18K & 102K & \cmark & \xmark & \xmark \\
nuScenes-QA~\cite{qian2023nuscenes} & 2023 & \cmark & \xmark & 340K & 460K & \cmark & \cmark & \xmark \\
NuPrompt~\cite{wu2023language} & 2023 & \xmark & \cmark & 34K & 35K & \cmark & \cmark & \xmark \\
DriveLM~\cite{drivelm2023} & 2023 & \cmark & \cmark & 34K & 375K & \cmark & \xmark & \xmark \\
MAPLM~\cite{tang2023thma,MAPLM} & 2023 & \cmark & \cmark & 2M & 16M & \cmark & \cmark & \cmark \\
\bottomrule
\end{tabular}
\caption{Multimodal-Language datasets for self-driving can be split to two types: (1) Added additional texts for existing nuScenes~\cite{caesar_nuscenes_2020} dataset such as Talk2Car~\cite{deruyttere_talk2car_2019}, nuScenes-QA~\cite{qian2023nuscenes}, DriveLM~\cite{drivelm2023}, and NuPrompt~\cite{wu2023language}; (2) independent collected datasets such as BDD-X~\cite{kim2018textual}, and DRAMA~\cite{malla2023drama}. }
\label{tab:vision-dataset}
\end{table*}


\section{LLVM-AD Workshop Summary}

The 1st \href{https://llvm-ad.github.io/}{LLVM-AD} is held together with WACV 2024 on Jan 8th, 2024 in Waikoloa, Hawaii. we seek to bring together academia and industry professionals in a collaborative exploration of applying MLLMs to autonomous driving. Through a half-day in-person event, the workshop will showcase regular and demo paper presentations and invited talks from famous researchers in academia and industry. Additionally, LLVM-AD will launch two open-source real-world traffic language understanding datasets, catalyzing practical advancements. The workshop will host two challenges based on this dataset to assess the capabilities of language and computer vision models in addressing autonomous driving challenges. 

\subsection{Multimodal Large Language Models for Autonomous Driving Challenges}

\paragraph{MAPLM Dataset.} Tencent's THMA HD Map AI labeling system is utilized to create descriptive paragraphs from HD map labels, offering nuanced portrayals of traffic scenes~\cite{tang2023thma}. Participants worked with various data modalities, including 2D camera images, 3D point clouds, and Bird's Eye View (BEV) images, enhancing our understanding of the environment. This innovative initiative explores the intersection of computer vision, AI-driven mapping, and natural language processing, highlighting the transformative potential of Tencent's THMA technology in reshaping our understanding and navigation of our surroundings.

\paragraph{UCU Dataset.} The primary objective of this challenge is the development of algorithms that are proficient in understanding drivers' commands and instructions represented as natural language input. These commands and instructions could encompass a diverse array of command types, ranging from safety-oriented instructions such as ``engage the emergency brakes" or ``adjust headlight brightness", to driving operational instructions such as ``shift to park mode" or ``set the cruise control to 70 mph", and comfort-related requests such as ``turn up the AC" or ``turn off seat heating". The scope of commands can even be extended to vehicle-specific instructions like ``open sunroof" or ``enable ego mode".

\subsection{Workshop Summary}

Nine papers were accepted in the inaugural Workshop on Large Language and Vision Models for Autonomous Driving (\href{https://llvm-ad.github.io/}{LLVM-AD}) at the 2024 IEEE/CVF Winter Conference on Applications of Computer Vision (WACV). They cover topics on MLLMs for autonomous driving focusing on integrating LLMs into user-vehicle interaction, motion planning, and vehicle control. Several papers explored the novel use of LLMs to enhance human-like interaction and decision-making in autonomous vehicles. For example, ``Drive as You Speak"~\cite{cui_2024_WACV} and ``Drive Like a Human"~\cite{fu_2024_WACV} presented frameworks where LLMs interpret and reason in complex driving scenarios, mimicking human behavior. ``Human-Centric Autonomous Systems With LLMs"~\cite{yang_2024_WACV} emphasized the importance of user-centric design, utilizing LLMs to interpret user commands. This approach represents a significant shift towards more intuitive and human-centric autonomous systems.

In addition to LLM integration, the workshop featured methodologies in vision-based systems and data processing. ``A Safer Vision-based Autonomous Planning System for Quadrotor UAVs"~\cite{zhong_2024_WACV} and ``VLAAD"~\cite{VALAAD_2024_WACV} demonstrated advanced approaches to object detection and trajectory planning, enhancing the safety and efficiency of UAVs and autonomous vehicles.

Optimizing technical processes was also a significant focus. For instance, ``A Game of Bundle Adjustment"~\cite{belder2023game} introduced a novel approach to improving 3D reconstruction efficiency, while ``Latency Driven Spatially Sparse Optimization"~\cite{Zampokas_2024_WACV} and ``LIP-Loc"~\cite{Shubodh_2024_WACV} explored advancements in CNN optimization and cross-modal localization, respectively. These contributions represent notable progress towards more efficient and accurate computational models in autonomous systems.

Furthermore, the workshop presented innovative approaches to data handling and evaluation. For example, NuScenes-MQA~\cite{Inoue_2024_WACV} introduced a dataset annotation technique for autonomous driving. Collectively, these papers illustrate a significant stride in integrating language models and advanced technologies into autonomous systems, paving the way for more intuitive, efficient, and human-centric autonomous vehicles.

\section{Discussion}

\paragraph{New Datasets for Multimodal Large Language Models in Autonomous Driving.}

Despite the success of LLMs in language understanding, applying them to autonomous driving presents a unique challenge. This is due to the necessity for these models to integrate and interpret inputs from diverse modalities, such as panoramic images, 3D point clouds, and HD map annotations. The current limitations in data scale and quality mean that existing datasets struggle to address all these challenges comprehensively. Furthermore, almost all multimodal LLMs like GPT-4V~\cite{gpt4v2023} have been pre-trained on a wealth of open-source datasets including traffic and driving scenes, the visual-language datasets annotated from nuScenes may not provide a robust benchmark for visual-language understanding in driving scene. Consequently, there is an urgent need for new, large-scale datasets that encompass a wide range of traffic and driving scenarios, including numerous corner cases, to effectively test and enhance these models in autonomous driving applications.

\paragraph{Hardware Support for Large Language Models in Autonomous Driving.}

In the use case of LLMs as the planner for autonomous driving, the perception reasoning for the LLMs and the subsequent control decision should be generated in real-time with low latency in order to meet safety requirements for autonomous driving. The number of (Floating-point operations per second)FLOPs of the LLMs has a positive correlation with the latency as well as the power consumption, which should be of consideration if LLMs are hosted in the vehicle. For LLMs deployed remotely, the bandwidth of perception information and control decision transfer will be a great challenge.

Another use case for LLMs in autonomous driving is a navigation planner~\cite{shah2023navigation,rajvanshi2023saynav}. Unlike driving planners, the tolerance of response time for the LLMs is much higher, and the number of queries for navigation planners is far less in general. Consequently, the hardware performance demand is easier to meet, and even moving the host to remote servers is a reasonable proposal.  

The user-vehicle interaction could also be a use case of LLMs in autonomous driving~\cite{cui_2024_WACV}. LLMs could interpret drivers' intentions into control commands given to the vehicle. For intentions unrelated to driving, e.g. entertainment control, the high latency of the response from LLMs could be accepted. However, if the intentions involve taking over autonomous driving, then the hardware requirements would be similar to the counterpart of using LLMs as an autonomous driving planner where LLMs are expected to respond with low latency.

LLMs in the applications of autonomous driving could potentially be compressed, which reduces the computation power requirements and the latency and lowers the HW limitation. However, the current effort in this field is still undeveloped.


\paragraph{Using Large Language Models for Understanding HD Maps.}

HD maps play a crucial role in autonomous vehicle technology, as they provide essential information about the physical environment in which the vehicle operates. The semantic map layer from the HD map is of utmost importance as it captures the meaning and context of the physical surroundings. To effectively encode this valuable information into the LLMs-powered next-generation autonomous driving, it is important to find a way to represent and comprehend the details of the environment in the language space.

Inspired by transformer-based language models, Tesla proposes a special language that they developed for encoding lanes and their connectivities. In this language of lanes, the words and tokens represent the lane positions in 3D space. The ordering of the tokens and predicted modifiers in the tokens encode the connectivity relationships between these lanes. Producing a lane graph from the model output sentence requires less post-processing than parsing a segmentation mask or a heatmap~\cite{tesla-ai-day-2022}. Pre-trained models (PTMs) have become a fundamental backbone for downstream tasks in natural language processing and computer vision. Baidu Maps has developed a system called ERNIE-GeoL, which has already been deployed in production. This system applies generic PTMs to geo-related tasks at Baidu Maps since April 2021, resulting in significant performance improvements for various downstream tasks~\cite{huang2022ernie}.

Tencent has developed an HD Map AI  system called THMA which is an innovative end-to-end, AI-based, active learning HD map labeling system capable of producing and labeling HD maps with a scale of hundreds of thousands of kilometers~\cite{tang2023thma}~\cite{zheng2023hdmap}. To promote the development of this field, they proposed the MAPLM~\cite{MAPLM} dataset containing over 2 million frames of panoramic 2D images, 3D LiDAR point cloud, and context-based HD map annotations, and a new question-answer benchmark MAPLM-QA.

\paragraph{User-Vehicle Interaction with Large Language Models.} 

Non-verbal language interpretation is also an important aspect to consider for user-autonomy teaming. Driver distraction poses a critical road safety challenge, including all activities such as smartphone use, eating, and interacting with passengers that divert attention from driving. According to the National Highway Traffic Safety Administration (NHTSA), distractions were a factor in 8.1\% of the 38,824 vehicle-related fatalities in the U.S. in 2020~\cite{stewart_overview_2022}. This issue becomes more pressing as semi-autonomous driving systems, particularly SAE Level 3 systems, gain prominence, requiring drivers to be ready to take control when prompted~\cite{sae_on-road_automated_vehicle_standards_committee_and_others_taxonomy_2018}.

To detect and mitigate driver distraction, driver action recognition strategies are commonly employed. These strategies involve continuous monitoring using sensors like RGB and infrared cameras, coupled with deep learning algorithms to identify and classify driver actions. Significant advancements have been made in this field~\cite{xing_driver_2019,ou_enhancing_2020,biparva_video_2022,ma_m2dar_2023,ma_cemformer_2023}.

Assessing the driver's cognitive state is also crucial, as it greatly indicates distraction levels. Physiological monitoring, such as through EEG signals, can provide insights into a driver's cognitive state~\cite{wang_online_2015,wang_eeg-based_2015}, but the intrusiveness of such sensors and their impact on regular driving patterns must be taken into account. Besides, behavior monitoring works such as through facial analysis, gaze, human pose, and motion~\cite{cao2023vitasd,lai2022eye,lai2023eye,lai2023listen,ijcai2022p0700,lee2021video} can also be used to analyze driver's driving status. Furthermore, current datasets on driver action recognition often lack mental state annotations required to train models in recognizing these states from sensory data, highlighting the need for semi-supervised learning methods to address this relatively unexplored challenge~\cite{ma_vit-dd_2023}.

\paragraph{Personlized Autonomous Driving.}


The integration of LLMs into autonomous vehicles marks a paradigm shift characterized by continuous learning and personalized engagement. LLMs can continuously learn from new data and interactions, adapting to changing driving patterns, user preferences, and evolving road conditions. This adaptability results in a refined and increasingly adept performance over time. Moreover, LLMs have the capability to be precisely fine-tuned or in-context learned to match individual driver preferences, furnishing personalized assistance that significantly improves the driving experience. This personalized approach enriches the driving experience, providing assistance that not only offers information but also aligns closely with the distinct requirements and subtleties of each driver.

Recent studies \cite{cui_2024_WACV,cui_receive_2023} have indicated the potential for LLMs to enhance real-time personalization in driving simulations, demonstrating their capacity to adapt driving behaviors in response to spoken commands. As the LLM-based personalization in autonomous driving is not well-developed, there are numerous opportunities for further research. Most recent studies focus on utilizing LLMs in the simulation environment instead of real vehicles. Integrating LLMs into actual vehicles is an exciting area of potential, moving beyond simulations to affect real-world driving experiences. Additionally, future investigations could also explore the development of LLM-driven virtual assistants that align with drivers' individual preferences, the employment of LLMs for the enhancement of safety features like fatigue detection, the application of these models in predictive vehicle maintenance, and the personalization of routing to align with drivers' unique inclinations. Furthermore, LLMs have the potential for personalizing in-vehicle entertainment, learning from drivers' behaviors to improve the driving experience.

\paragraph{Trustworthy and Safety for Autonomous Driving.}

Another crucial takeaway is enhancing transparency and trust. When the vehicle makes a complex decision, such as overtaking another vehicle on a high-speed, two-lane highway, passengers and drivers might naturally have questions or concerns. In these instances, the LLM doesn't just execute the task but also articulates the reasoning behind each step of the decision-making process. By providing real-time, detailed explanations in understandable language, the LLM demystifies the vehicle's actions and underlying logic. This not only satisfies the innate human curiosity about how autonomous systems work but also builds a higher level of trust between the vehicle and its occupants.

Moreover, the advantage of ``zero-shotting" was particularly evident during the complex overtaking maneuver on a high-speed Indiana highway. Despite the LLM not having encountered this specific set of circumstances before—varying speeds, distances, and even driver alertness—it was able to use its generalized training to safely and efficiently generate a trajectory for the overtaking action. With some uncertainty estimation techniques~\cite{miao2021prevent, ye_mitigating_2023,xiong2023can}, this can ensure that even in dynamic or edge case scenarios, the system can make sound judgments while keeping the user informed, therefore building confidence in autonomous technology.

To sum up, LLMs demonstrate their potential to revolutionize autonomous driving by enhancing safety, transparency, and user experience. Tasked with complex commands like overtaking, the LLM considered real-time data from multiple vehicle modules to make informed decisions, clearly articulating these to the driver. The model also leveraged its zero-shot learning capabilities to adapt to new scenarios, providing personalized, real-time feedback. Overall, the LLM proved effective in building user trust and improving decision-making in autonomous vehicles, emphasizing its utility in future automotive technologies.

\section{Conclusion}
\label{sec:concl}

 In this survey, we explored the pattern of integrating multimodal large language models (MLLMs) into the next generation of autonomous driving systems. Our study began with an overview of the development of both MLLMs and autonomous driving, which have traditionally been considered distinct fields but are now increasingly interconnected. Then, we conducted an extensive literature review on the specific algorithms and applications of multimodal language models for autonomous driving and then focused on the current state of research and benchmarking datasets that apply MLLMs to autonomous driving. A significant highlight of our study was the synthesis of key insights and findings from the first \href{https://llvm-ad.github.io/}{LLVM-AD} workshop such as proposing new datasets and improving current MLLMs algorithms on autonomous driving. Finally, we engaged in a forward-looking discussion on vital research themes and the promising potential for enhancing MLLMs in autonomous driving. We discussed both challenges and opportunities that lie ahead, aiming to show the pathway for further exploration. In general, this paper serves as a valuable resource for researchers in the autonomous driving area. It offers a comprehensive understanding of the significant role and vast potential that MLLMs hold in revolutionizing the landscape of autonomous transportation. We hope this paper could facilitate research in integrating MLLMs with autonomous driving in the future.

\section*{Acknowledgments}
We would like to express our gratitude for the support received from the Purdue University Digital Twin Lab (\href{https://purduedigitaltwin.github.io/}{https://purduedigitaltwin.github.io/}), Tencent T Lab, and PediaMed AI (\href{https://pediamedai.com/}{http://pediamedai.github.io/}) for their contributions to this survey paper.

{\small
\bibliographystyle{ieee_fullname}
\bibliography{egbib, bib/can, bib/juanwu, bib/references, bib/zichong,bib/wscite}
}

\end{document}